\def\year{2022}\relax
\documentclass[letterpaper]{article} 
\usepackage{aaai22}  
\usepackage{times}  
\usepackage{helvet}  
\usepackage{courier}  
\usepackage[hyphens]{url}  
\usepackage{graphicx} 
\usepackage{natbib}  
\usepackage{caption} 
\DeclareCaptionStyle{ruled}{labelfont=normalfont,labelsep=colon,strut=off} 
\usepackage{algorithm}
\usepackage{algorithmicx} 
\usepackage[noend]{algpseudocode} 

\usepackage{newfloat}
\usepackage{listings}
\floatstyle{ruled}
\newfloat{listing}{tb}{lst}{}
\floatname{listing}{Listing}
\title{Isconna: Streaming Anomaly Detection with Frequency and Patterns}
\author{
	Rui Liu,
	Siddharth Bhatia,
	Bryan Hooi
}
\affiliations{

	National University of Singapore \\
	xxliuruiabc@gmail.com, \{siddharth, bhooi\}@comp.nus.edu.sg
}

\usepackage{mathtools} 
\usepackage{booktabs} 
\usepackage{subcaption}
\usepackage{numprint}
\usepackage[utf8]{inputenc}
\usepackage{etoolbox}
\usepackage{tikz}
\usepackage{mathrsfs} 
\usepackage{amsfonts} 
\usepackage{amssymb} 
\usepackage{amsmath} 
\usepackage{soul} 
\usepackage{diagbox}

\newtheorem{definition}{Definition}
\newtheorem{problem}{Problem}
\newtheorem{theorem}{Theorem}

\begin{document}

	\maketitle

	\begin{abstract}
		An edge stream is a common form of presentation of dynamic networks.
		It can evolve with time, with new types of nodes or edges being continually added.
		Existing methods for anomaly detection rely on edge occurrence counts or compare pattern snippets found in historical records.
		In this work, we propose Isconna, which focuses on both the frequency and the pattern of edge records.
		The burst detection component targets anomalies between individual timestamps, while the pattern detection component highlights anomalies across segments of timestamps.
		These two components together produce three intermediate scores, which are aggregated into the final anomaly score.
		Isconna does not actively explore or maintain pattern snippets;
		it instead measures the consecutive presence and absence of edge records.
		Isconna is an online algorithm, it does not keep the original information of edge records;
		only statistical values are maintained in a few count-min sketches (CMS).
		Isconna's space complexity $O(rc)$ is determined by two user-specific parameters, the size of CMSs. In worst case, Isconna's time complexity can be up to $O(rc)$, but it can be amortized in practice.
		Experiments show that Isconna outperforms five state-of-the-art frequency- and/or pattern-based baselines on six real-world datasets with up to 20 million edge records.
	\end{abstract}

	\section{Introduction}

	An edge stream, or a stream of edge records, represents a set of connections on a time-evolving graph.
	Unlike its static counterpart, in a stream, edge records reach the detection system one after another.
	Therefore, some statistical values for the whole dataset are not available as a priori;
	this causes problems in data preprocessing and other aspects.

	In the real world, an edge stream can be used as the abstraction of various applications, such as network connections, social network posts.
	Within these streams, anomalies may occasionally appear, like distributed denial of service (DDoS) attacks in a network graph, breaking events in a social network graph, or money laundering in a transaction network.
	As an informal definition, anomalies in an edge stream are behaviors, represented as edges, that deviates from the usual pattern or are known to be harmful to the system's regular operation.
	The goal of anomaly detection algorithms is to mark those anomalous behaviors, where a typical method is to assign a score to each edge record indicating the degree of anomalousness.

	Existing methods, like SedanSpot, process each edge record individually without considering the contextual information.
	If edge records are extracted outside of the original context, the algorithm tends to produce the same result.
	However, in different environments, what should be regarded as anomalous are significantly different.
	The normal level of an industry network is unlikely to be the same as a home network.
	While recent methods take into account contextual information, they only focus on a specific attribute;
	for example, MIDAS uses the occurrence count for burst detection.
	In this work, as a preliminary attempt, we combine the characteristics of multiple contextual attributes, proposing the Isconna, a streaming anomaly detection algorithm focusing on both robustness and efficacy.

	\begin{figure}[!htb]
		\centering
		\includegraphics[width=0.49\linewidth]{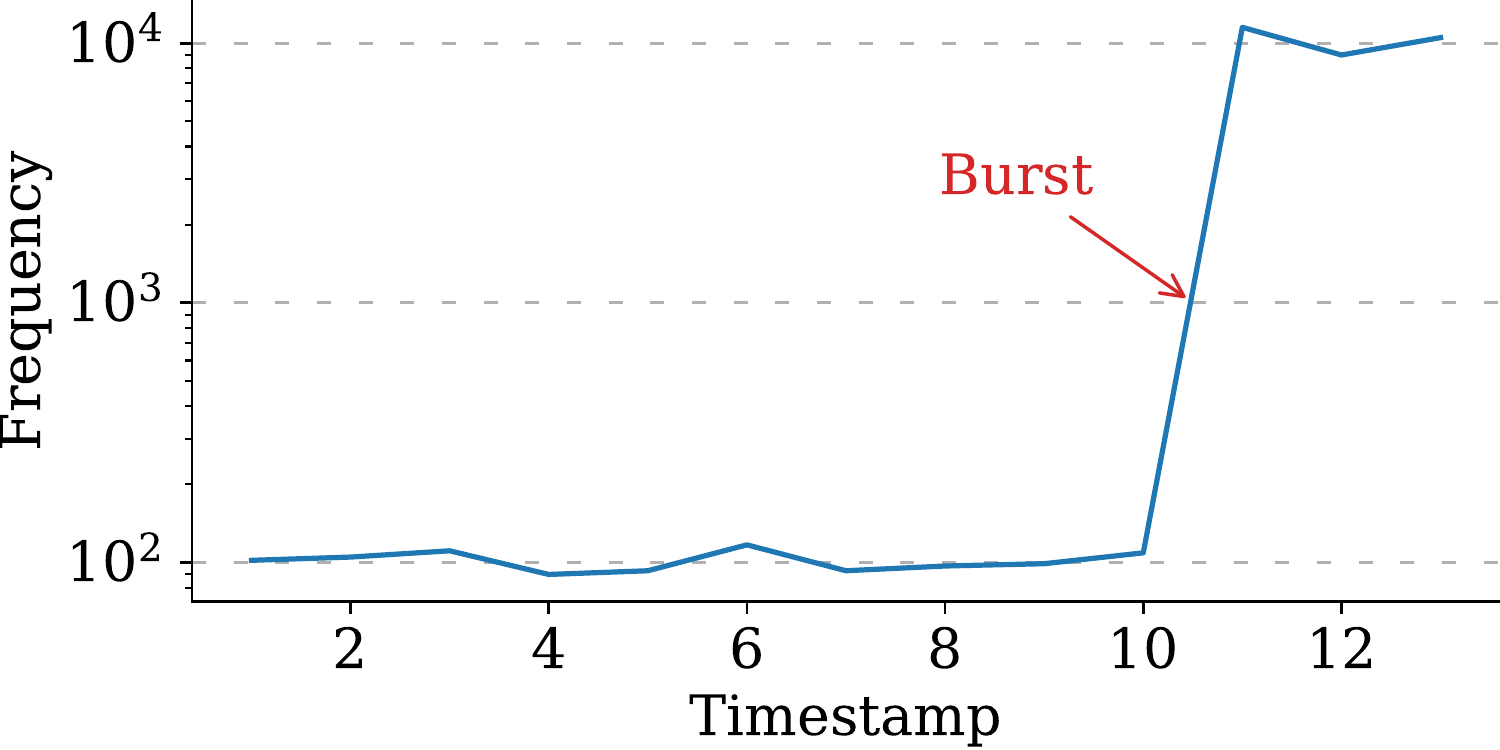}
		\hfill
		\includegraphics[width=0.49\linewidth]{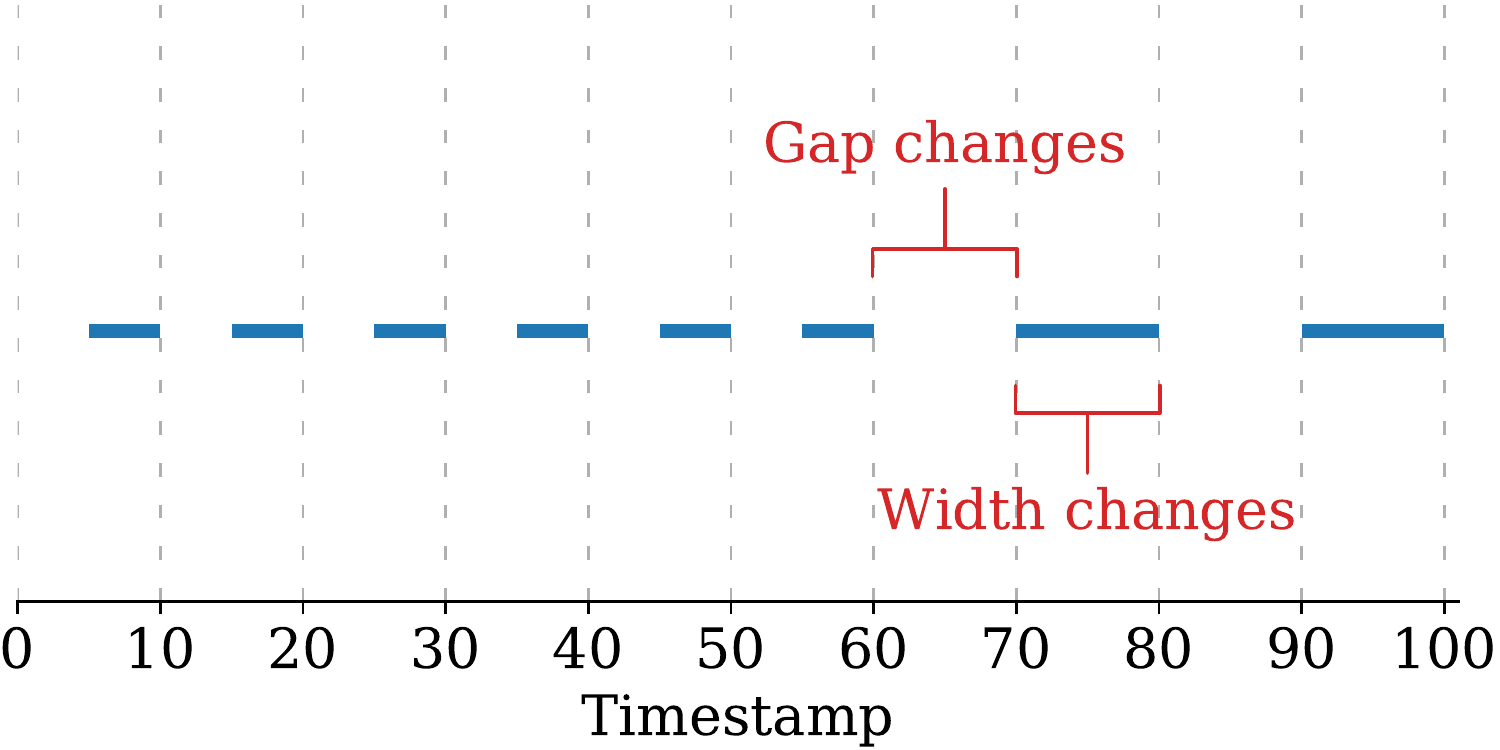}
		\caption{Example of bursts (left) and pattern changes (right)}\label{fig:Example}
	\end{figure}

	Informally, a burst is a sudden change in the number of occurrences (frequency).

	As an example, Fig.~\ref{fig:Example} (left) shows the number of outbound connections from a workstation within consecutive timestamps.
	Before timestamp 11, there are around 100 connections in each timestamp.
	From timestamp 11, it surges to about 10,000 and maintains in the following timestamps.
	Intuitively, we can believe something happens to the machine, e.g., the user account is compromised, or the network adapter is malfunctioning.

	Likewise, the change of pattern may also indicate irregular events.
	Fig.~\ref{fig:Example} (right) gives an example.
	It represents whether a user visits a news website, and each timestamp is one day.
	Before day 60, the user visits the website for five consecutive days, then does not visit it for another five days, and loop.
	After day 60, the length of the consecutive visiting and non-visiting are both doubled.
	This potentially indicates some event occurs and changes the user's reading habit.

	Our algorithm is motivated by the burst and the pattern change.
	It learns the frequency, the consecutive occurrence and absence from the history data, then uses them to predict the anomalous degree of future data.

	Existing methods in the streaming anomaly detection are briefly introduced in the next section.
	The details of the Isconna is given in the Proposed Algorithm section.
	The Experiments section demonstrates the performance of Isconna using various experiments with different conditions and configurations.

	\section{Related Works}

	Anomaly Detection is a vast topic by itself and cannot be fully covered in this manuscript.
	See \cite{Chandola2009Anomaly} for an extensive survey on anomaly detection, and \cite{Akoglu2015Graph} for a survey on graph-based anomaly detection.

	OddBall \cite{Akoglu2010oddball}, CatchSync \cite{Jiang2016Catching} and \cite{Kleinberg1999Authoritative} detect anomalous nodes.
	AutoPart \cite{Chakrabarti2004AutoPart} spots anomalies by finding edge removals that significantly reduce the compression cost.
	NrMF \cite{Tong2011Non} detect anomalous edges by factoring the adjacency matrix and flagging edges with high reconstruction errors.
	FRAUDAR \cite{Hooi2017Graph} and k-cores \cite{Shin2018Patterns} target anomalous subgraphs detection,
	However, these approaches work on static graphs only.

	Among methods that focus on dynamic graphs, DTA/STA \cite{Sun2006streams} approximate the adjacency matrix of the current snapshot using the matrix factorization.
	AnomRank \cite{Yoon2019Fast}, which is inspired by PageRank \cite{Page1999PageRank}, iteratively updates two score vectors and compute anomaly scores.
	Copycatch \cite{Beutel2013CopyCatch} and SpotLight \cite{Eswaran2018SpotLight} detect anomalous subgraphs.
	HotSpot \cite{Yu2013Anomalous}, IncGM+ \cite{Abdelhamid2017Incremental} and DenseAlert \cite{Shin2017DenseAlert} utilize incremental method to process graph updates or subgraphs more efficiently.
	SPOT/DSPOT \cite{Siffer2017Anomaly} use the extreme value theory to automatically set thresholds for anomalies.
	However, these approaches work on graph snapshots only and are unable to handle the finer granularity of edge streams.

	As for methods focusing on edge streams, RHSS \cite{Ranshous2016Scalable} focuses on sparsely-connected parts of a graph.
	SedanSpot \cite{Eswaran2018SedanSpot} identifies edge anomalies based on edge occurrence, preferential attachment, and mutual neighbors.
	MIDAS \cite{Bhatia2020Midas} identifies microcluster-based anomalies, or suddenly arriving groups of suspiciously similar edges.
	CAD \cite{Sricharan2014Localizing} localizes anomalous changes in the graph structure using the commute time distance measurement.
	However, these methods are unable to detect any seasonality or periodic patterns in the data.

	PENminer \cite{Belth2020Mining} explores the persistence of activity snippets, i.e., the length and regularity of edge-update sequences' reoccurrences.
	F-FADE \cite{Chang2021F} aims to detect anomalous interaction patterns by factorizing the frequency of those patterns.
	These methods can effectively detect periodic patterns, but they require a considerable amount of time to explore and maintain possible patterns in the network.

	Recently, several deep learning based methods have also been proposed for anomaly detection;
	see \cite{Chalapathy2019Deep} for an extensive survey.
	However, such approaches are unable to detect in a streaming manner.

	\section{Theory}

	\subsection{Table of Notations}

	Due to the page limit, the table of notations is given in the supplementary.

	\subsection{Definitions}

	\subsubsection{Edge Record}

	An edge record $e$ is an individual entry in the dataset.
	It is defined as an ordered tuple $(s,d,t)$.
	The array of all edge records sorted in non-descending order of timestamp is called an edge stream $\mathscr S$.

	\subsubsection{Edge Type}

	An edge type $e$ is an abstract edge without the timestamp $t$.
	It is defined as an ordered tuple $(s,d)$.
	The set of all edge types is the edge set $\mathscr E$.

	\subsubsection{Segment}

	A segment $g$, or $[t_i,t_j]$ is an set of consecutive timestamps, defined as $\{t_k|t_i\le t_k\le t_j\}$.
	For a segment $[t_i,t_j]$, if there is an edge type $e\coloneqq(s,d)$ additionally satisfies $\forall t\in[t_i,t_j], \exists e\in\mathscr S, s=s_e\land d=d_e\land t=t_e$, then the segment is called an \textbf{occurrence segment} of edge type $e$ on the interval $t_i$ to $t_j$, or an occurrence segment.
	Similarly, a segment is called an \textbf{absence segment} iff $\forall t\in[t_i,t_j], \nexists e\in\mathscr S, s=s_e\land d=d_e\land t=t_e$.
	In this work, we only consider non-expandable segments, that is, $[t_i,t_j]$ is an occurrence segment or an absence segment, but neither $[t_i-1,t_j]$ nor $[t_i,t_j+1]$ is.
	We additionally define $g(e)$ as the occurrence segment that contains edge record $e$ and $g(e,t)$ as the absence segment of edge type $e$ with timestamp $t$ included.

	\begin{problem}[Anomaly Detection]
		Given an edge stream $\mathscr S$ on a time-evolving graph $G$, assign a score to each edge record $e$ according to the contextual information, where a higher score indicates a higher degree of anomalousness.
	\end{problem}

	\subsection{Proposed Anomalousness Measure}

	An anomalousness measure is a function that evaluates the anomalousness of a given edge record according to its internal states.
	Formally, $A(e;\theta)\coloneqq\mathscr S\times\Theta\to \mathbb R^+$, where $\theta\in\Theta$ is the contextual information maintained as the internal state of the detector, $\Theta$ is the set of all possible internal states, $A(\cdot)$ is the anomalousness measure.
	We first propose the combined form of the anomalousness measure
	\begin{equation}
		\label{eq:c5ed84bb-3dc3-406f-82bf-043d4d122929}
		A(e;\theta)\coloneqq f(B(e;\theta),P(e;\theta))
	\end{equation}
	where $B(\cdot)$ is a function that evaluates the burst anomalousness, $P(\cdot)$ is a function that evaluates pattern change anomalousness, and $f(\cdot)$ is a function that combines these components.

	\subsubsection{Burst Anomalousness Measure}

	For burst detection, we adopt the idea of the MIDAS algorithm, using a statistical hypothesis test to evaluate the degree of deviation from the learned mode.
	For simplicity, we derive the formula of anomalousness from the G-test.
	The comparison between the chi-squared test and the G-test is out of the scope of this work, a detailed discussion can be found at \cite{Cressie1984Multinomial}.

	\begin{align*}
		G & \coloneqq2\sum_i O_i\cdot\log\dfrac{O_i}{E_i} \\
		& =2\left(O_\text{current}\cdot\log\dfrac{O_\text{current}}{E_\text{current}}+O_\text{past}\cdot\log\dfrac{O_\text{past}}{E_\text{past}}\right) \\
		& =2\left[c\cdot\log\dfrac{c}{\dfrac{a}{t-1}}+a\cdot\log\dfrac{a}{\dfrac{a}{t-1}\cdot(t-1)}\right] \\
		& =2c\cdot\log\left[\dfrac{c\cdot(t-1)}{a}\right]
	\end{align*}
	where $G$ is the G-test statistic, $O_i$ is the observed count of class $i$, $E_i$ is the expected count of class $i$, $t$ is the timestamp of edge record $e$, $c$ is the current occurrence count of $e$ in timestamp $t$, $a$ is the accumulated occurrence count of $e$ up to timestamp $t-1$.

	The derived statistic ranges from negative infinity to positive infinity.
	However, as an anomalousness measure, we only concern the extent of deviations.
	Hence, we take its absolute value.
	The formal definition of the burst anomalousness measure is given in Definition~\ref{def:BurstAnomalousnessMeasure}.

	\begin{definition}[Burst Anomalousness Measure]
		\label{def:BurstAnomalousnessMeasure}
		Given an edge record $e\in\mathscr S$ and the detector's internal state $\theta$, the burst anomalousness measure is defined as
		\begin{align}
			\label{eq:B(e;theta)}
			B(e;\theta)\coloneqq & \left|2c(e;\theta)\cdot\log\left[\dfrac{c(e;\theta)(t_e-1)}{a(e;\theta)}\right]\right| \\
			c(e;\theta)\coloneqq & \|\{e_i|e=e_i,e_i\in\mathscr S,t_{e_i}=t_\theta\}\| \\
			a(e;\theta)\coloneqq & \|\{e_i|e=e_i,e_i\in\mathscr S,t_{e_i}<t_\theta\}\|
		\end{align}
		where $B(e;\theta)$ is the burst anomalousness measure, $t_e$ is the timestamp of edge record $e$, $t_\theta$ is detector's internal timestamp.
	\end{definition}

	\subsubsection{Pattern Change Anomalousness Measure}

	For pattern change detection, an intuitive idea is to maintain a portion of the history, then detect the repeated pattern.
	If we regard each edge record as a character and the maintained history as a string, this becomes the repeated substring pattern problem.
	For example, string ``abcabcabcabc'' is four repeats of substring ``abc''.
	There are many different algorithms to solve this problem efficiently.
	However, as string comparison is unavoidable, it would be difficult, if not impossible, to decrease the time complexity below linear.
	Additionally, with the edge stream keeps evolving, the maintained records are also constantly updated.
	In the worst case, the above algorithm for detecting repeated patterns is executed whenever a new edge record reaches the system.
	On the other hand, the maintained stream history may be a combination of multiple patterns with different periods.
	As a result, such an exact pattern tracking solution may not be practical.

	We try to use various methods to solve or bypass those problems.
	The first method is to track individual edge types rather than exact patterns.
	In theory, if there are $n$ edge types and the maximum pattern length is $m$, then the number of unique patterns would be $O(n^m)$.
	Therefore, targeting edge types can significantly reduce the space complexity, and thus the time complexity.

	The second method is to abstract patterns into several statistical values.
	We choose the length of patterns and the interval between pattern occurrences because they are easier to maintain.
	For simplicity, we call them width and gap, respectively.
	Note that we do not use the term ``period'', because in the real world, patterns do not always have a strict period, but may have a rough range of intervals between occurrences.
	This method alone does not help solve the aforementioned problems.
	However, with different edge types being tracked separately, updating the maintained history is equivalent to modifying the recorded width and gap values of affected edge types.

	As a quick summary, two statistical values, the length of consecutive edge type occurrences (width) and the interval between occurrences (gap), are used to compute the pattern change anomalousness, whose formula is not discussed yet.

	For the formula of pattern change anomalousness measure, we still prefer the statistical hypothesis tests due to their high interpretability.
	To achieve this, we split the tracking of width and gap into the current and the historical values, and use the concept of segments to differentiate them.
	Table~\ref{tab:93065655-b870-4ce5-8327-ebfeb60250e8} shows those combinations and corresponding notations that will be used in the definition.
	The formal definition of the pattern change anomalousness measure is given in Definition~\ref{def:PatternChangeAnomalousnessMeasure}.

	\begin{table}[!htb]
		\centering
		\caption{Notations of concept combinations}
		\label{tab:93065655-b870-4ce5-8327-ebfeb60250e8}
		\begin{tabular}{ccc}
			\toprule
			& Current & Accumulated \\
			\midrule
			Occurrence & $c(g(e);\theta)$ & $a(g(e);\theta)$ \\
			\midrule
			Absence & $c(g(e,t);\theta)$ & $a(g(e,t);\theta)$ \\
			\bottomrule
		\end{tabular}
	\end{table}

	\begin{definition}[Pattern Change Anomalousness Measure]
		\label{def:PatternChangeAnomalousnessMeasure}
		Given an edge record $e\in\mathscr S$ and the detector's internal state $\theta$, the pattern change anomalousness measure is defined as
		\begin{align}
			\label{eq:P(e;theta)}
			P(e;\theta)\coloneqq & f(P(g(e);\theta),P(g(e,t');\theta)) \\
			\label{eq:P(g;theta)}
			P(g;\theta)\coloneqq & \left|2c(g;\theta)\cdot\log\left[\dfrac{c(g;\theta)(\bar t(g)-1)}{a(g;\theta)}\right]\right| \\
			c(g;\theta)\coloneqq & \|\{t|t\in g\}\| \\
			a(g(e);\theta)\coloneqq & \sum_{[t_i,t_j]\in\mathscr G_o\backslash g(e),t_j<t_e}\|\{t|t\in[t_i,t_j]\}\| \\
			a(g(e,t);\theta)\coloneqq & \sum_{[t_i,t_j]\in\mathscr G_a\backslash g(e,t),t_j<t}\|\{t|t\in[t_i,t_j]\}\|
		\end{align}
		where $P(e;\theta)$ is the pattern change anomalousness measure, $t'$ is the largest timestamp that is smaller than the initial timestamp of $g(e)$, $\bar t(g)$ is the index number of $g$ \footnotemark, $\mathscr G_o$ is the set of occurrence segments, $\mathscr G_a$ is the set of absence segments.
	\end{definition}

	\footnotetext{Occurrence segments and absence segments are counted separately.}

	\subsubsection{Component Combinator}

	In Equation~\ref{eq:c5ed84bb-3dc3-406f-82bf-043d4d122929} and Equation~\ref{eq:P(e;theta)}, we use a combinator $f(\cdot)$ without giving its definition.
	Since the components defined above represent different aspects of the stream, they may not have the same scale.
	Thus we define $f(\cdot)$ as the product of components with exponential weights.
	The anomalousness measure can be therefore converted into the following formula.

	\begin{equation}
		\label{eq:A(e;theta)}
		A(e;\theta)\coloneqq B(e;\theta)^\alpha\cdot P(g(e);\theta)^\beta\cdot P(g(e,t');\theta)^\gamma
	\end{equation}
	where the exponents $\alpha,\beta,\gamma\in[0,+\infty)$ are the weight of components, depending on the goals of the actual scenario.

	As a general guide, $\beta$ is usually greater than $\gamma$, because an occurrence segment includes the information about this edge type, while an absence segment is about other edge types and is already covered by their corresponding occurrence segments.
	For $\alpha$ vs. $\beta$, it depends on the characteristics of the edge stream.
	For example, in a stable scenario where fluctuations are relatively small, a higher $\alpha$ helps spot bursts that are difficult to notice.
	While in the scenario where edges follow strong patterns, users may prefer a high $\beta$ and $\gamma$ value to capture any abnormal changes.

	\section{Proposed Method: Isconna}

\begin{figure*}[!htb]
		\centering
		\includegraphics[width=0.9\textwidth]{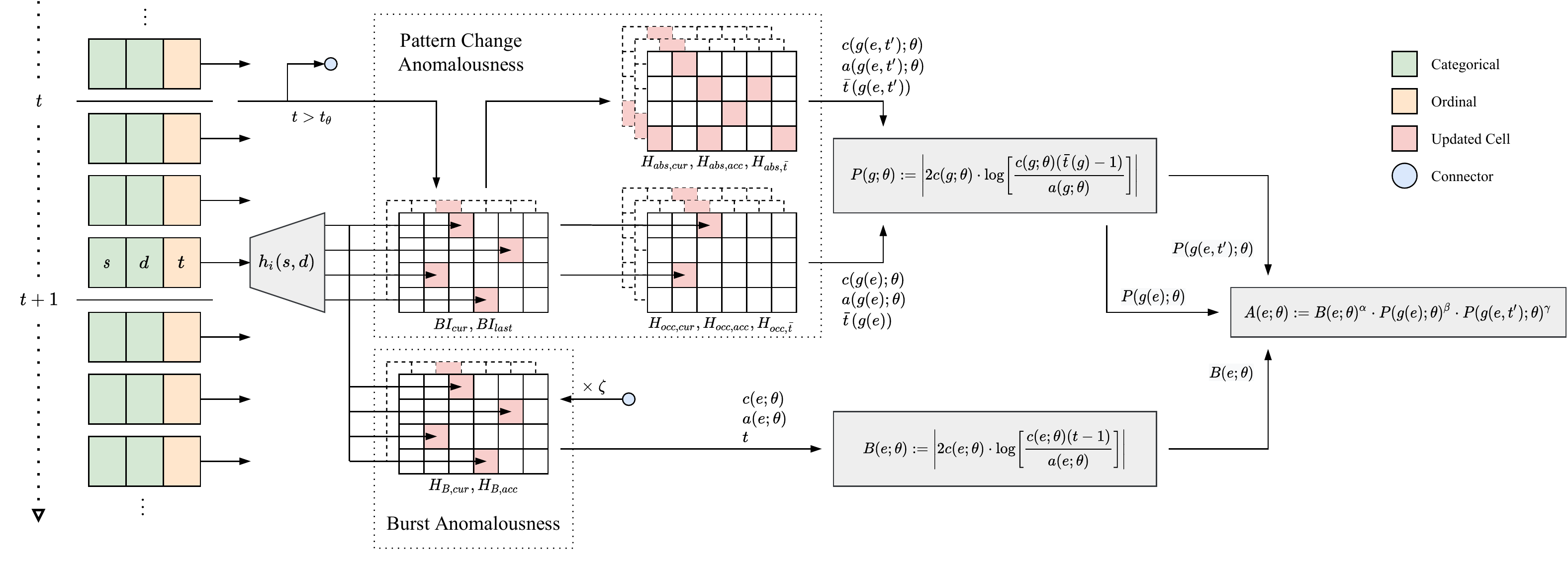}
		\caption{Overview of the Isconna algorithm}\label{fig:Overview}
	\end{figure*}

	In this section, we will describe the algorithm procedure and implementation details.
	Fig.~\ref{fig:Overview} gives an overview of our proposed Isconna algorithm.


	\subsection{Count-Min Sketches}

	From Equation~\ref{eq:A(e;theta)}, we learn that for each edge record, the detector needs to compute the anomalousness measure for the three different components.
	As edges are tracked by edge types, ideally, the detector should maintain the information of each edge type independently.
	However, this is impractical due to the memory limit and the streaming nature.

	To resolve this problem, we use \textbf{Count-Min Sketches} (CMS) \cite{Cormode2004Improved} to store the intermediate information.
	A CMS consists of a few hash tables with different hash functions but the same number of cells.
	In addition to the Hash, Add and Query operations provided by the original CMS, we also define the ArgQuery operation.
	A brief introduction to CMS is given in the supplementary.

	For simplicity, we apply the \textbf{Same-Layout Assumption} to CMSs in the detector, that is, all CMSs share the same shape and the same group of hash functions, thus an item can be hashed into the same group of cells for all CMSs.
	In practice, this assumption reduces the number of required hashing operations and helps decrease the running time.

	\subsection{Busy Indicators}

	The tracking for the burst anomalousness measure is straightforward.
	As it only counts the number of occurrences, when an edge reaches the detector, it is hashed and the corresponding cells in CMSs increment by 1.
	However, for the pattern change anomalousness measure, it tracks the length of occurrence/absence segments, which requires special data structures to track the beginning and the ending of segments.
	Therefore, we propose a special type of CMS, \textbf{Busy Indicators} (BI).

	A group of BIs consists of two CMSs with Boolean elements, one tracks the edge occurrence of the current timestamp, another for the last timestamp.
	All possible element combinations are listed in Table~\ref{tab:7400afb4-5814-4461-a9d1-22511a506391}.
	Note that the beginning of an occurrence segment is at the same time the ending of the adjacent absence segment, and vice versa.

	\begin{table}[!htb]
	    \centering
	    \caption{Combination of BI element values}\label{tab:7400afb4-5814-4461-a9d1-22511a506391}
	    \begin{tabular}{ccc}
			\toprule
			\diagbox{Cur.}{Last} & $\top(>0)$ & $\bot(=0)$ \\
			\midrule
			$\top(>0)$ & Occ. cont. & Occ. init. \\
			\midrule
			$\bot(=0)$ & Abs. init. & Abs. cont. \\
			\bottomrule
	    \end{tabular}
	\end{table}

	\subsection{Procedure}

	\begin{algorithm}[!htb]
		\caption{Isconna-EO ($\mathscr S$, $\zeta$, $\alpha$, $\beta$, $\gamma$)}\label{alg:Isconna-EO}
		\begin{algorithmic}[1]
			\Require Stream $\mathscr S$, scale factor $\zeta$, weight $\alpha$, $\beta$, $\gamma$
			\Ensure Anomalousness measure $A$
			\State Initialize CMS $H_{B,cur}$, $H_{B,acc}$, $H_{occ,cur}$, $H_{occ,acc}$, $H_{occ,\bar t}$, $H_{abs,cur}$, $H_{abs,acc}$, $H_{abs,\bar t}$, $BI_{cur}$ and $BI_{last}$
			\State Initialize internal state $\theta$ \Comment{Including CMSs}
			\For{new edge record $e\in\mathscr S$}
				\If{$t_\theta<t_e$} \label{ln:Isconna-EO.EndOfTimestamp} \label{ln:fae90800-5d04-4bcb-b4f8-55dda50f5f85}
					\State \Call{Absence}{$\theta$, $\zeta$}
					\State $H_{B,cur}\gets \zeta\cdot H_{B,cur}$
					\State $t_\theta\gets t_e$
				\EndIf \label{ln:85fab417-00f5-4257-a532-3a2b16a3f110}
				\State $I\gets$ \Call{Hash}{$H_{B,cur}$, $e$} \label{ln:818b48d4-75db-403f-b7ad-e7e15e5766cd}
				\State \Call{Add}{$H_{B,cur}$, $I$}
				\State \Call{Add}{$H_{B,acc}$, $I$}
				\State $c(e;\theta)\gets$ \Call{Query}{$H_{B,cur}$, $I$}
				\State $a(e;\theta)\gets$ \Call{Query}{$H_{B,acc}$, $I$}
				\State Compute $B(e;\theta)$ according to Equation~\ref{eq:B(e;theta)} \label{ln:4494acfc-c582-4ae2-ae3a-1257558a8f38}
				\State \Call{Occurrence}{$\theta$, $\zeta$, $I$} \label{ln:b5f47c3f-56cf-4b69-a00a-920b7e94a179}
				\State $i\gets$ \Call{ArgQuery}{$H_{occ,\bar t}$, $I$} \label{ln:87c0ad31-e578-450a-aaf9-50419cc39054}
				\State $c(g(e);\theta)\gets H_{occ,cur}(i)$
				\State $a(g(e);\theta)\gets H_{occ,acc}(i)$
				\State $\bar t(g(e))\gets H_{occ,\bar t}(i)$
				\State Compute $P(g(e);\theta)$ according to Equation~\ref{eq:P(g;theta)}
				\State $i\gets$ \Call{ArgQuery}{$H_{abs,\bar t}$, $I$} \label{ln:e2d97938-00fc-4c05-9730-8a34ea9af20b}
				\State $c(g(e,t');\theta)\gets H_{abs,cur}(i)$
				\State $a(g(e,t');\theta)\gets H_{abs,acc}(i)$
				\State $\bar t(g(e,t'))\gets H_{abs,\bar t}(i)$
				\State Compute $P(g(e,t');\theta)$ according to Equation~\ref{eq:P(g;theta)} \label{ln:dfcc72c0-0214-46c2-a1bc-d72f9920032e}
				\State \textbf{yield} $A(e;\theta)$ according to Equation~\ref{eq:A(e;theta)} \label{ln:c46c5b0b-8c6f-45af-af7f-3fe70ba96686}
			\EndFor
		\end{algorithmic}
	\end{algorithm}

	\begin{algorithm}[!htb]
		\caption{Functions for Pattern Change Anomalousness}\label{alg:AdditionalFunctions}
		\begin{algorithmic}[1]
			\Function{Occurrence}{$\theta$, $\zeta$, $I$} \label{ln:b83bf13f-a759-4a37-8215-6fab99b8e49e}
				\For{index $i\in I$}
					\If{$\neg BI_{cur}(i)$}
						\State $BI_{cur}(i)\gets\top$ \label{ln:UpdateWidth.BcGetTrue}
						\If{$\neg BI_{last}(i)$}
							\State $H_{occ,acc}(i)\gets H_{occ,acc}(i)+H_{occ,cur}(i)$
							\State $H_{occ,cur}(i)\gets \zeta\cdot H_{occ,cur}(i)$
							\State $H_{occ,\bar t}(i)\gets H_{occ,\bar t}(i)+1$
						\EndIf
						\State $H_{occ,cur}(i)\gets H_{occ,cur}(i)+1$
					\EndIf
				\EndFor
			\EndFunction \label{ln:a1286666-6173-4317-9f87-d7bbef89cf48}
			\Function{Absence}{$\theta$, $\zeta$} \label{ln:9dcf9e6a-05a4-4c3d-8673-ab9870eee95f}
				\For{$i\in 0:\Call{sizeof}{H_{abs,cur}}$}
					\If{$\neg BI_{cur}(i)$}
						\If{$BI_{last}(i)$}
							\State $H_{abs,acc}(i)\gets H_{abs,acc}(i)+H_{abs,cur}(i)$
							\State $H_{abs,cur}(i)\gets \zeta\cdot H_{abs,cur}(i)$
							\State $H_{abs,\bar t}(i)\gets H_{abs,\bar t}(i)+1$
						\EndIf
						\State $H_{abs,cur}(i)\gets H_{abs,cur}(i)+1$
					\EndIf
					\State $BI_{last}(i)\gets BI_{cur}(i)$
					\State $BI_{cur}(i)\gets\bot$
				\EndFor
			\EndFunction \label{ln:9148a33e-b05b-4ed1-a7df-745e4e05fdd3}
		\end{algorithmic}
	\end{algorithm}

	Algorithm~\ref{alg:Isconna-EO} gives the pseudocode of the Isconna algorithm.
	The high-level structure of the algorithm includes two stages: regular processing (line \ref{ln:818b48d4-75db-403f-b7ad-e7e15e5766cd}-\ref{ln:c46c5b0b-8c6f-45af-af7f-3fe70ba96686}) and additional steps when the timestamp advances (line~\ref{ln:fae90800-5d04-4bcb-b4f8-55dda50f5f85}-\ref{ln:85fab417-00f5-4257-a532-3a2b16a3f110}).

	Line~\ref{ln:818b48d4-75db-403f-b7ad-e7e15e5766cd}-\ref{ln:4494acfc-c582-4ae2-ae3a-1257558a8f38} are for the burst anomalousness measure.
	Due to the same-layout assumption, we only need to hash once for the cell indices, they can be reused on all other CMSs.

	Line~\ref{ln:b5f47c3f-56cf-4b69-a00a-920b7e94a179}-\ref{ln:dfcc72c0-0214-46c2-a1bc-d72f9920032e} are for the pattern change anomalousness measure.
	It first calls function Occurrence to update the information of occurrence segment $g(e)$.
	Function Occurrence (Algorithm~\ref{alg:AdditionalFunctions}, line \ref{ln:b83bf13f-a759-4a37-8215-6fab99b8e49e}-\ref{ln:a1286666-6173-4317-9f87-d7bbef89cf48}) first checks if this edge type is already processed in the current timestamp.
	If not, it checks whether $e$ is the beginning of a new occurrence segment, and performs merging and resetting steps.
	Line~\ref{ln:87c0ad31-e578-450a-aaf9-50419cc39054} and line~\ref{ln:e2d97938-00fc-4c05-9730-8a34ea9af20b} call function ArgQuery function to retrieve the cell index of the minimum segment index number.
	Since counts in CMSs are no less than their actual value, the minimum segment index is the fewest updated one, and thus the most accurate one.

	Line~\ref{ln:fae90800-5d04-4bcb-b4f8-55dda50f5f85}-\ref{ln:85fab417-00f5-4257-a532-3a2b16a3f110} are the additional steps when the timestamp advances.
	Function Absence (Algorithm~\ref{alg:AdditionalFunctions}, line~\ref{ln:9dcf9e6a-05a4-4c3d-8673-ab9870eee95f}-\ref{ln:9148a33e-b05b-4ed1-a7df-745e4e05fdd3}) resembles function Occurrence.
	However, since it is called at the end of timestamps, it iterates over all cells and additionally resets BIs.

	Like the predecessor MIDAS algorithm, we take into account the temporal effect, thus the algorithm requires an additional parameter $\zeta$, the scale factor.
	The value of $\zeta$ depends on the density of edge records in each timestamp.
	As an empirical guide, if there are many edge records in each timestamp or the same edge type is expected to reoccur within a few timestamps, then a low $\zeta$ is preferred.
	For example, in a wide area network (WAN), each timestamp is one second, there will be more than millions of connections per second, and the connection between the same pair of nodes is highly likely to reoccur in the next second.
	On the contrary, if edge records are sparse in the stream, a high $\zeta$ maintains the information for a longer time and prevents too many near-empty cells in the CMSs.

	\subsection{Edge-Node (EN) Variant}

	Apart from the edge-only (EO) version as shown in Algorithm~\ref{alg:Isconna-EO}, we also propose an edge-node (EN) variant which additionally incorporates the node information.
	Due to the space limit, the pseudocode of the Isconna-EN algorithm is given in the supplementary.

	Processing nodes are similar to processing edge records, except that function Hash takes one argument instead.
	For simplicity, we replace the intermediate steps with three instances of Isconna-EO, but note that those sub-instances return the anomalousness measure of each component and do have a for loop inside.

	Compared with edge records, there are fewer nodes in a graph.
	This has several advantages:
	(1) The CMS conflict occurs less often;
	(2) The average length of occurrence segments is higher, and that of absence segments is lower, which makes width scores and gap scores more balanced.

	However, this does not necessarily mean the EN variant is always better than the original EO variant.
	Although extra information provides a more comprehensive understanding of the graph, it may also interfere with the algorithm's decision making.

	\subsection{Theoretical Guarantee}

	Our theoretical guarantee establishes a bound on the false positive probability of the detection.

	\begin{theorem}
		Let $\chi^2_{1-\delta}(1)$ be the $1-\delta$ quantile of a $\chi^2$ random variable with 1 degree of freedom.
		Then
		\begin{equation}
			P(\tilde G>\chi^2_{1-\delta}(1))<\delta
		\end{equation}
		where $\delta$ is a parameter of the CMS;
		$\tilde G$ is the adjusted G-test statistic.
	\end{theorem}

	In other words, if $\tilde G$ is the test statistic and $\chi^2_{1-\delta}(1)$ is the threshold, the probability of producing a false positive, i.e., $\tilde G$ is incorrectly higher than $\chi^2_{1-\delta}(1)$, is at most $\delta$.
	Details and the proof are given in the supplementary.

	\subsection{Time and Space Complexity}\label{sec:Complexity}

	For the space complexity, the algorithm does not keep the original edge information, only count/length values in CMSs are maintained.
	As the CMS is the only data structure used in the algorithm, if we denote $r$ as the number of rows of a CMS, $c$ as the number of columns of a CMS, then Isconna's space complexity is $O(rc)$.

	For the time complexity, the end-of-timestamp processing (the if block in Alg.~\ref{alg:Isconna-EO}) and the rest part are discussed separately.
	In the end-of-timestamp processing, function Absence iterates over all the cells in $H_{abs,cur}$.
	If we denote each CMS has $r$ rows and $c$ columns, the time complexity of this part is $O(rc)$.
	For the rest processing steps (after the if block in Alg.~\ref{alg:Isconna-EO}), all the CMS operations and the UpdateWidth function only visit one cell in each row;
	hence the time complexity is $O(r)$.
	However, we cannot simply combine the time complexity of both parts, as the number of edge records in each timestamp is unknown.
	In the worst case, where there is only one edge record in each timestamp, the overall time complexity is $O(rc)$ for each edge record, although this is unlikely in practice.

	\section{Experiments}\label{sec:Experiments}

	\subsubsection{Datasets}

	We use six real-world datasets, CIC-DdoS2019~\cite{Sharafaldin2019Developing}, CIC-IDS2018~\cite{Sharafaldin2018Generating}, CTU-13~\cite{Garcia2014empirical}, DARPA~\cite{Lippmann1999Results}, ISCX-IDX2012~\cite{Shiravi2012developing} and UNSW-NB15~\cite{Moustafa2015UNSW}. Due to the space limit, the statistical details are given in the supplementary.

	\subsubsection{Baselines}

	We use SedanSpot, PENminer, F-Fade, and MIDAS as baselines.
	All the algorithms have their open-source implementation provided by the authors, where SedanSpot and MIDAS are implemented in C++, PENminer, and F-Fade are implemented in Python.
	Apart from the default parameter provided in the source code, we further tested some other combinations of parameters.
	The detailed parameter combinations are listed in the supplementary.

	\subsubsection{Evaluation Metrics}

	All methods output an anomaly score for each input edge record (higher is more anomalous).
	The area under the receiver operating characteristic curve (AUROC) is reported as the accuracy metric.
	For the speed metric, the running time of the program (excluding I/O) is used as the metric.
	Unless explicitly specified, all experiments including baselines are repeated 11 times and the median is reported.

	\subsubsection{Experimental Setup}

	We implement Isconna in C++.
	Experiments are performed on a Windows 10 machine with a 4.20GHz Intel i7-7700K CPU and 32GiB RAM.

	\subsection{Performance}\label{sec:Experiment.Accuracy}

	\begin{table*}[!htb]
		\centering
		\caption{AUROC of each method on different datasets}\label{tab:Experiment.AUROC}
		\makebox[\textwidth][c]{
			\begin{tabular}{lrrrrrrr}
				\toprule
				Dataset      & PENminer & F-Fade & SedanSpot & MIDAS  & MIDAS-R & \textbf{Isconna-EO} & \textbf{Isconna-EN} \\
				\midrule
				CIC-DDoS2019 & $---$    & 0.7802 & 0.5893    & 0.6746 & 0.9918  & \textbf{0.9995}     & 0.9987              \\
				CIC-IDS2018  & 0.8209   & 0.6179 & 0.4834    & 0.5516 & 0.9240  & \textbf{0.9975}     & 0.9758              \\
				CTU-13       & 0.6041   & 0.8028 & 0.6435    & 0.8900 & 0.9730  & \textbf{0.9827}     & 0.9477              \\
				DARPA        & 0.8724   & 0.9283 & 0.7119    & 0.8780 & 0.9494  & 0.9323              & \textbf{0.9678}     \\
				ISCX-IDS2012 & 0.5300   & 0.6797 & 0.5948    & 0.3823 & 0.7728  & 0.9250              & \textbf{0.9654}     \\
				UNSW-NB15    & 0.7028   & 0.6859 & 0.8435    & 0.8841 & 0.8928  & \textbf{0.9195}     & 0.8974              \\
				\bottomrule
			\end{tabular}
		}
	\end{table*}

	Isconna produces higher accuracy (AUROC) while being fast in comparison to the baselines.

	Table~\ref{tab:Experiment.AUROC} shows the AUROC of baselines and our proposed algorithms on the 6 real-world datasets.
	Parameter values that produce the highest AUROC on each dataset are reported in the supplementary.
	Note that the PENminer algorithm on CIC-DDoS2019 cannot finish within 24 hours;
	thus, this result is not reported in Table~\ref{tab:Experiment.AUROC}.
	Our proposed Isconna outperforms SedanSpot, PENminer, F-Fade, and MIDAS on all datasets.

	\begin{figure*}[!htb]
		\centering
		\begin{minipage}{0.335\textwidth}
			\includegraphics[width=\textwidth,height=\textheight,keepaspectratio]{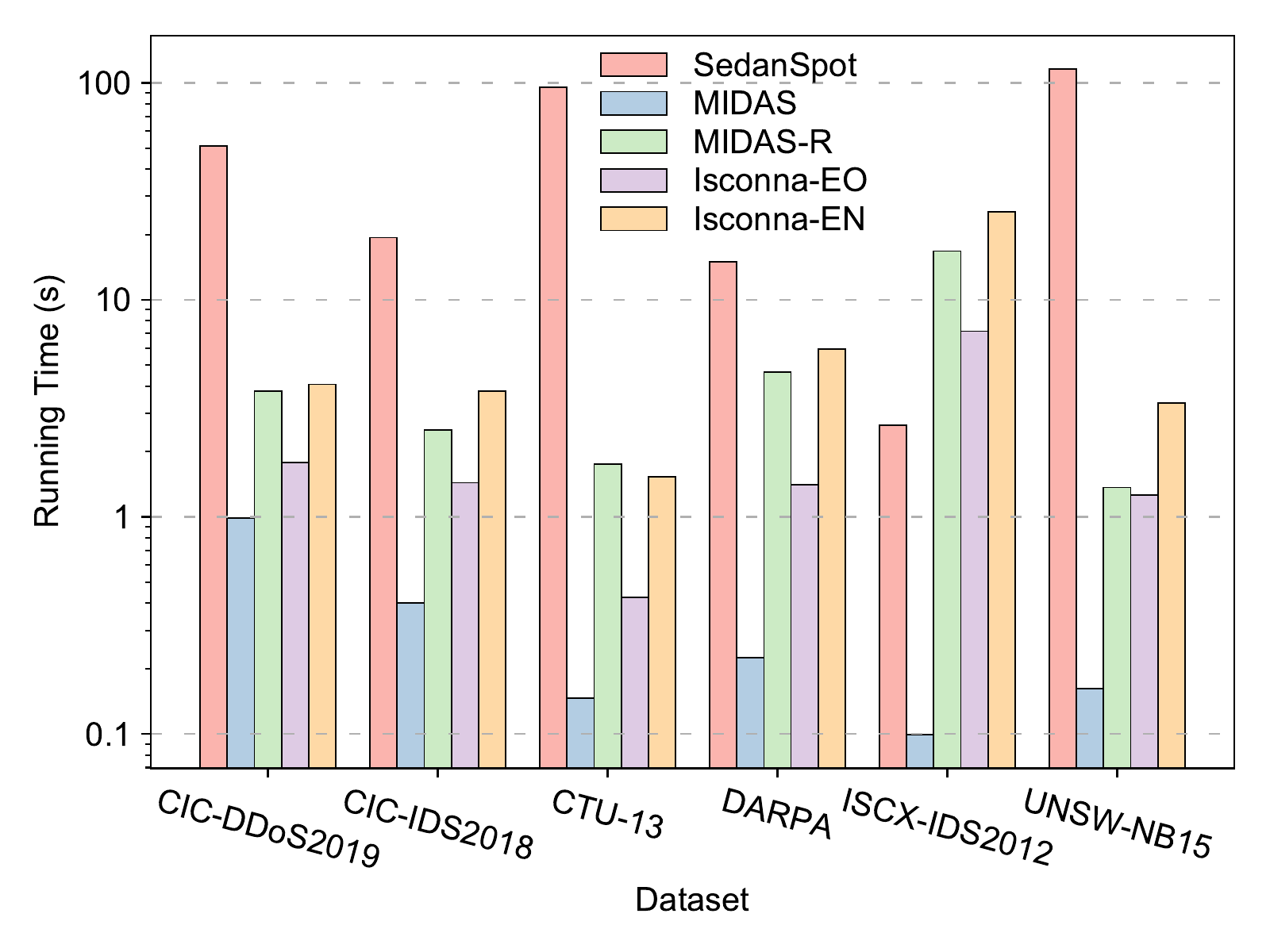}
			\caption{Time cost of Isconna and baselines}\label{fig:Experiment.Time}
		\end{minipage}%
		\hfill
		\begin{minipage}{0.635\textwidth}
			\begin{subfigure}{0.49\textwidth}
				\includegraphics[width=\textwidth,height=\textheight,keepaspectratio]{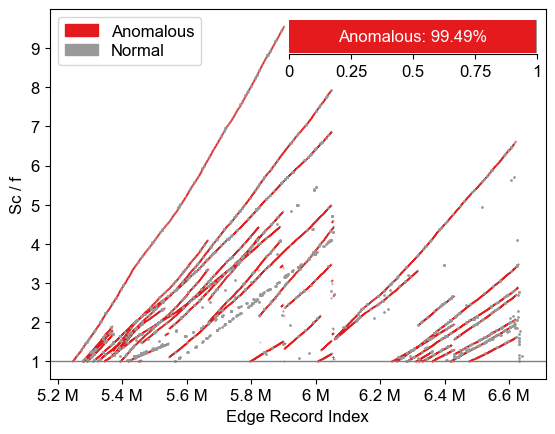}
			\end{subfigure}%
			\hfill
			\begin{subfigure}{0.51\textwidth}
				\includegraphics[width=\textwidth,height=\textheight,keepaspectratio]{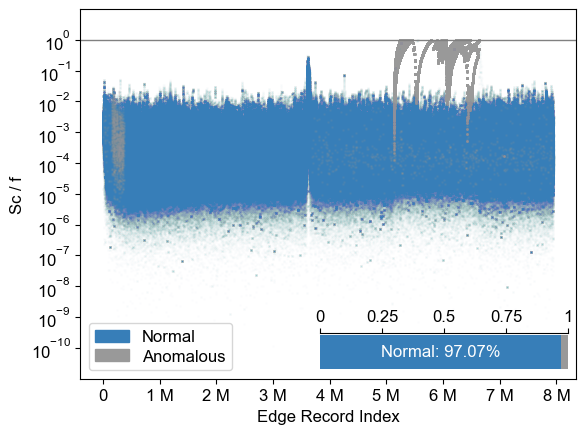}
			\end{subfigure}%
			\caption{The contribution of pattern change detection to the final anomalousness measure, for final score greater than frequency score (left) and less than (right)}
			\label{fig:Experiment.Qualitative.Positive}
			\label{fig:Experiment.Qualitative.Negative}
		\end{minipage}
	\end{figure*}

	Fig.~\ref{fig:Experiment.Time} shows the speed comparison of SedanSpot, MIDAS, MIDAS-R, Iscoona-EO, and Isconna-EN algorithms.
	F-Fade and PENminer are omitted since they are implemented in Python and spend minutes to hours processing large datasets (orders of magnitude slower).
	Note that on datasets other than ISCX-IDS2012, SedanSpot is significantly slower than other algorithms.
	On the other hand, MIDAS is the fastest among all algorithms.
	However, as seen in Table~\ref{tab:Experiment.AUROC}, this fast speed is at the cost of low detection accuracy.


	\subsection{Effectiveness of Pattern Detection}

	\begin{figure*}[!htb]
		\centering
		\begin{subfigure}{0.32\textwidth}
			\centering
			\includegraphics[width=\textwidth,height=\textheight,keepaspectratio]{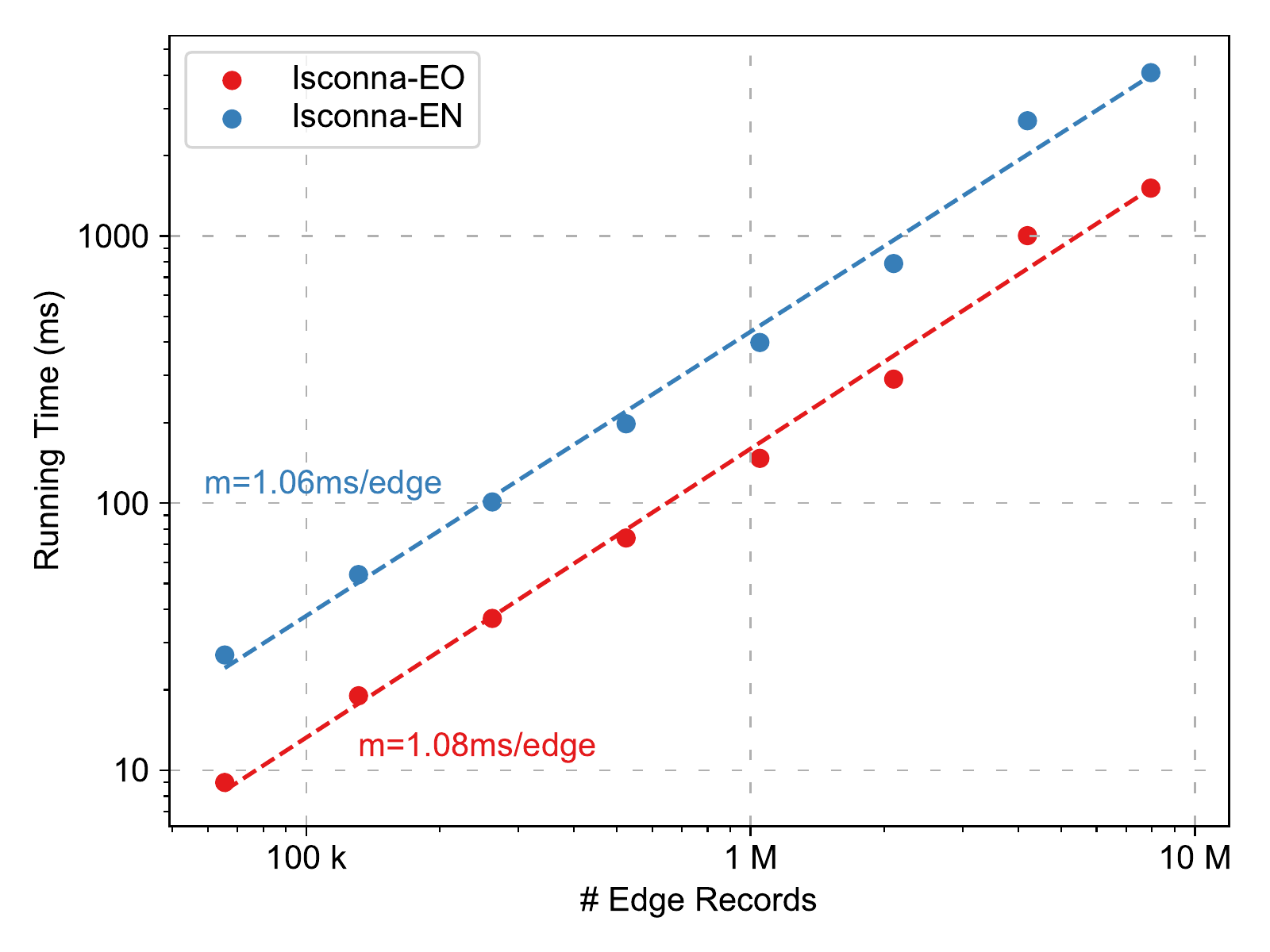}
			\caption{For the number of edge records}\label{fig:Experiment.Scalability.Edge}
		\end{subfigure}%
		\hfill
		\begin{subfigure}{0.32\textwidth}
			\centering
			\includegraphics[width=\textwidth,height=\textheight,keepaspectratio]{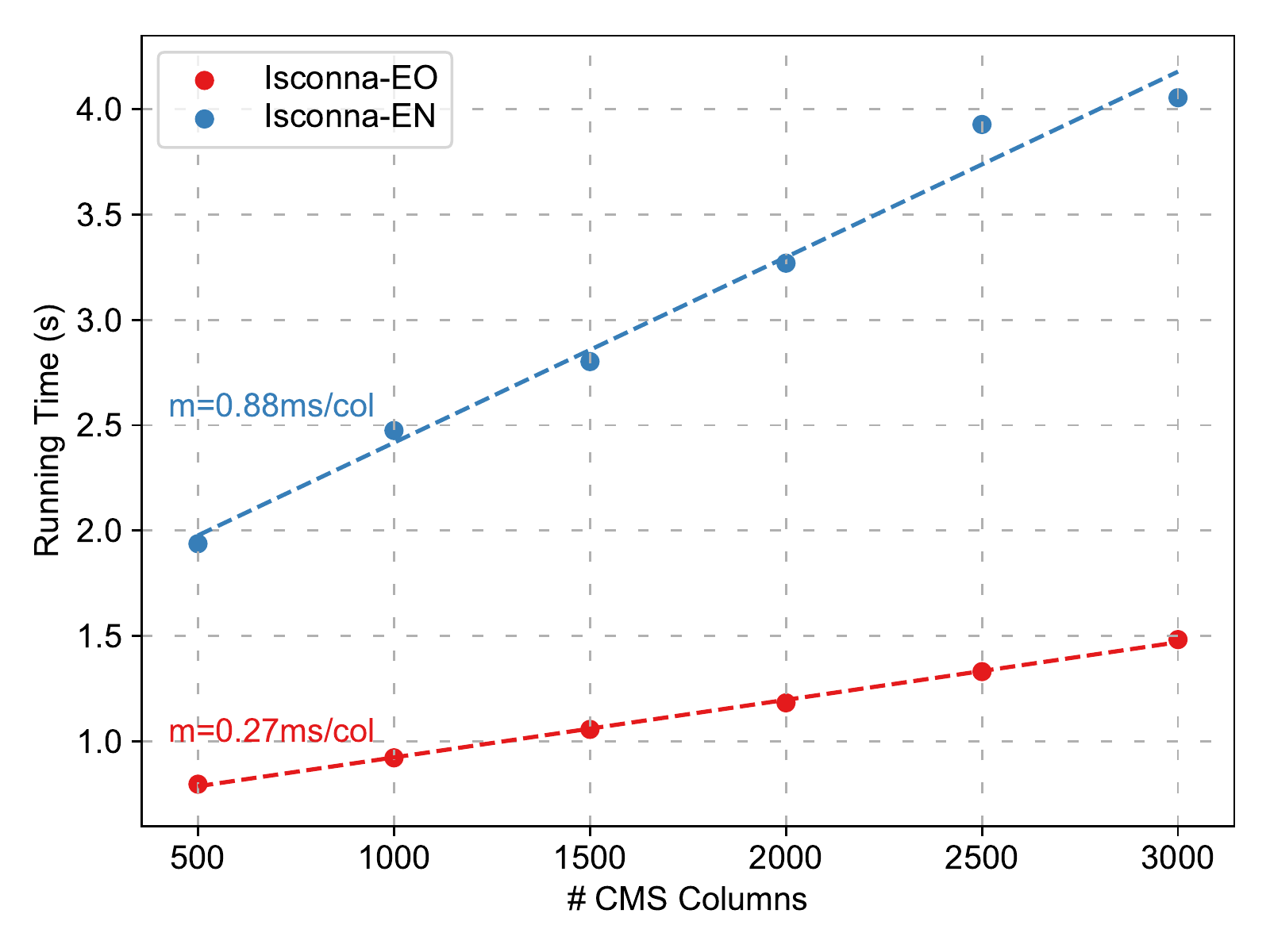}
			\caption{For the number of CMS columns}\label{fig:Experiment.Scalability.Column}
		\end{subfigure}%
		\hfill
		\begin{subfigure}{0.32\textwidth}
			\centering
			\includegraphics[width=\textwidth,height=\textheight,keepaspectratio]{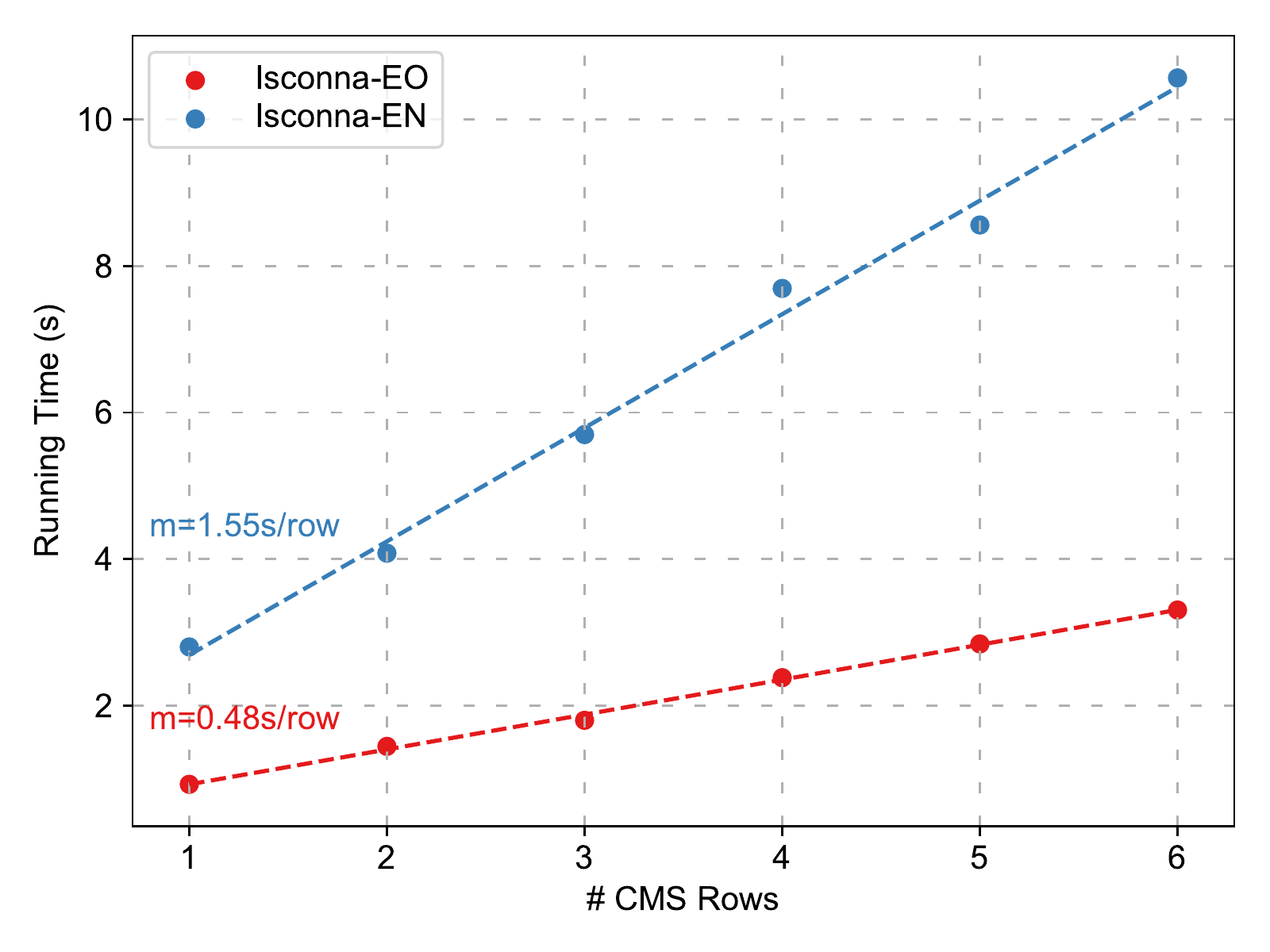}
			\caption{For the number of CMS rows}\label{fig:Experiment.Scalability.Row}
		\end{subfigure}%
		\caption{Scalability (time cost vs. different parameter values)}\label{fig:Experiment.Scalability}
	\end{figure*}

	In this part, we demonstrate the effectiveness of the pattern detection component by comparing the anomaly scores with and without width score and gap scores, i.e., comparing $A(e;\theta)$ and $B(e;\theta)$.
	We run Isconna-EO on the CIC-IDS2018 dataset.
	Fig.~\ref{fig:Experiment.Qualitative.Positive} (left) shows the edge records whose anomaly scores are increased after multiplying the pattern change score, i.e., $A(e;\theta)/B(e;\theta)>1$.
	Among them, 99.49\% are true positive and 0.51\% are false positive.
	This demonstrates that pattern detection can help better distinguish anomalous edge records by producing higher anomaly scores.
	On the other hand, Fig.~\ref{fig:Experiment.Qualitative.Negative} (right) shows edge records where $A(e;\theta)/B(e;\theta)<1$, of which 97.07\% are true negative and 2.93\% are false negative.
	This indicates that pattern detection is able to reduce the probability of incorrectly labeling normal edge records as anomalous, with an acceptable false negative rate.

	\subsection{Scalability}


	Fig.~\ref{fig:Experiment.Scalability.Edge} shows the scalability of edge records.
	We test the required time to process the first $2^{16},2^{17},\dots,2^{22}$ and all edge records of the CIC-IDS2018 dataset.
	The results are spread around two lines, which indicates the running time of the algorithm on the whole dataset is linear to the number of edge records in it.
	Thus we can confirm the constant time complexity for processing individual edge records.

	Fig.~\ref{fig:Experiment.Scalability.Column} and Fig.~\ref{fig:Experiment.Scalability.Row} show the CMS column scalability and the CMS row scalability, respectively.
	We can see the running time is linear to the number of columns/rows.
	This demonstrates that the size of CMSs is one of the terms of time complexity.
	Additionally, we may notice the slope of the EN variant is about three times of the EO variant.
	This is due to the extra CMSs for source nodes and destination nodes;
	both have a group of CMSs that resembles those of edges.

	\section{Conclusion}

	In this paper, we propose Isconna-EO and Isconna-EN algorithms, which detect burst and pattern anomalies in a streaming manner, without actively exploring or maintaining pattern snippets.
	The time complexity of processing an individual edge records is constant with respect to the data scale.
	Additionally, we provide a theoretical guarantee on the false positive probability.
	Our experimental results show that Isconna outperforms $5$ state-of-the-art frequency- or pattern-based baselines on $6$ real-world datasets, and demonstrate the scalability effectiveness of the pattern detection component.
	Future work could consider compatibility for general types of data and techniques for automatic parameter tuning.

	\bibliography{main}

\end{document}


\maketitle

	\appendix

	\section{Table of Notations}

	\begin{table}[!htb]
		\centering
		\caption{Notations}\label{tab:Notation}
		\begin{tabular}{cl}
			\toprule
			Notation   & Description                   \\
			\midrule
			$s$ & Source node (categorical) \\
			$d$ & Destination node (categorical) \\
			$t$ & Timestamp (ordinal) \\
			$e$ & Edge type $(s,d)$ or edge record $(s,d,t)$ \\
			$\mathscr E$ & Edge (type) set \\
			$\mathscr S$ & Edge (record) stream \\
			$g$, $[t_i,t_j]$ & Segment \\
			$g(e)$ & Occurrence segment \footnotemark \\
			$g(e,t)$ & Absence segment \footnotemark \\
			$\bar t(g)$ & Segment index number \\
			$t'$ & Timestamp before the beginning of $g(e)$ \\
			$\mathscr G$ & Segment set \\
			$A(\cdot)$ & Anomalousness measure \\
			$B(\cdot)$ & Burst anomalousness measure \\
			$P(\cdot)$ & Pattern change anomalousness measure \\
			$\theta$ & Algorithm internal state \\
			$c(e;\theta)$ & Current edge count \\
			$a(e;\theta)$ & Accumulated edge count \\
			$c(g;\theta)$ & Current segment length \\
			$a(g;\theta)$ & Accumulated segment length \\
			$\alpha$ & Weight for $B(\cdot)$ \\
			$\beta$  & Weight for $P(g(e);\cdot)$ \\
			$\gamma$ & Weight for $P(g(e,t');\cdot)$ \\
			$\zeta$ & Scale factor \\
			$H_{B,cur}$ & CMS for current edge counts \\
			$H_{B,acc}$ & CMS for accumulated edge counts \\
			$H_{occ,cur}$ & CMS for current occurrence segment length \\
			$H_{occ,acc}$ & CMS for accumulated occurrence segment length \\
			$H_{occ,\bar t}$ & CMS for occurrence segment index number \\
			$H_{abs,cur}$ & CMS for current absence segment length \\
			$H_{abs,acc}$ & CMS for accumulated absence segment length \\
			$H_{abs,\bar t}$ & CMS for absence segment index number \\
			$BI_{cur}$ & Busy indicator for current timestamp \\
			$BI_{last}$ & Busy indicator for last timestamp \\
			\bottomrule
		\end{tabular}
	\end{table}

	\footnotetext[1]{Parameter $e$ is the edge record.}
	\footnotetext[2]{Parameter $e$ is the edge type.}

	\section{Count-Min Sketch}

	\begin{figure}[!htb]
		\centering
		\begin{tikzpicture}[scale=0.75]
			\def\varIndex{{0,3,1,4,2}}
			\node (o) at (-4,-2) {Object};
			\foreach \i in {1,...,4} {
					\node (h\i) at (-1.5,-\i+0.5) {$h_\i(o)=I_\i$};
					\node (cell\i) at (0.5+\varIndex[\i],-\i+0.5) {$+1$};
					\draw[very thick] (\varIndex[\i],-\i) rectangle (\varIndex[\i]+1,-\i+1);
					\draw[->] (o) -- (h\i);
					\draw[->] (h\i) -- (cell\i);
				}
			\draw[step=1cm,gray,very thin] (0,0) grid (6,-4);
			\undef\varIndex
		\end{tikzpicture}
		\caption{Function Hash and Add}\label{fig:CMS.HashAdd}
	\end{figure}

	\begin{figure}[!htb]
		\centering
		\begin{tikzpicture}[scale=0.75]
			\def\varIndex{{0,3,1,4,2}}
			\node (min) at (9,-2) {$\min\{v_1,\ldots,v_4\}$};
			\foreach \i in {1,...,4} {
					\node (h\i) at (-1,-\i+0.5) {$I_\i$};
					\node (cell\i) at (0.5+\varIndex[\i],-\i+0.5) {$v_\i$};
					\draw[very thick] (\varIndex[\i],-\i) rectangle (\varIndex[\i]+1,-\i+1);
					\draw[->] (h\i) -- (cell\i);
					\draw[->] (cell\i) -- (6.5,-\i+0.5) -- (min);
				}
			\draw[step=1cm,gray,very thin] (0,0) grid (6,-4);
			\undef\varIndex
		\end{tikzpicture}
		\caption{Function Query}\label{fig:CMS.Query}
	\end{figure}

	We will briefly introduce the CMS operations used in this work.
	A detailed description can be found at \cite{Cormode2004Improved}.
	Figure~\ref{fig:CMS.HashAdd} illustrates the function Hash and Add.
	Function Hash runs the hash function of each CMS row on the given object to obtain the cell indices.
	Function Add takes those computed indices and adds 1 to those cells.
	Figure~\ref{fig:CMS.Query} illustrates the function Query.
	It takes the computed indices, then obtains the elements in those cells and return the minimum.
	Another function ArgQuery is similar to function Query, only that the final $\min$ is replaced with $\arg\min$, which returns the index of the minimal element.

	\section{Pseudocode of Isconna-EN}

	\begin{algorithm}[!htb]
		\caption{Isconna-EN}
		\label{alg:Isconna-EN}
		\begin{algorithmic}[1]
			\Require Edge stream $\mathscr S$, scale factor $\zeta$, weight $\alpha$, $\beta$, $\gamma$
			\Ensure Anomalousness measure $A$
			\State Initialize Isconna-EO instance $\mathit{Isc}$, $\mathit{Isc}'$ $\mathit{Isc}''$
			\For{new edge record $e\coloneqq(s,d,t)\in E$}
				\State $\{B(e;\theta),P(g(e);\theta),P(g(e,t');\theta)\}\gets\mathit{Isc}(e)$
				\State $\{B(s;\theta),P(g(s);\theta),P(g(s,t');\theta)\}\gets\mathit{Isc}'(s)$
				\State $\{B(d;\theta),P(g(d);\theta),P(g(d,t');\theta)\}\gets\mathit{Isc}''(d)$
				\State $B(e;\theta)\gets\max(B(e;\theta),B(s;\theta),B(d;\theta))$
				\State $P(g(e);\theta)\gets\max(P(g(e);\theta),P(g(s);\theta),P(g(d);\theta))$
				\State $P(g(e,t');\theta)\gets\max(P(g(e,t');\theta),P(g(s,t');\theta),P(g(d,t');\theta))$
				\State \textbf{yield} Compute $A(e;\theta)$
			\EndFor
		\end{algorithmic}
	\end{algorithm}

	\section{Theoretical Guarantee}\label{sec:Appendix.TheoreticalGuarantee}

	The result of Isconna is not a binary decision but an anomaly score.
	The upper bound of scores may be different on different datasets.
	Thus, it may be difficult to tell whether a score indicates an anomalous edge record.
	We try to alleviate this problem by providing a false positive probability to a user-defined threshold.
	In this section, we explicitly separate the actual value and its CMS approximation by adding hat symbol ( $\hat{}$ ) to the latter.

	According to \cite{Cormode2004Improved}, a CMS has two hyperparameters $\varepsilon$ and $\delta$.
	If we denote the number of rows as $d$ and the number of columns as $w$, we can obtain them from $\varepsilon$ and $\delta$ using equations below.

	\begin{align}
		w & =\left\lceil\dfrac{e}{\varepsilon}\right\rceil \label{eqn:TheoreticalGuarantee.CMS.Column} \\
		d & =\left\lceil\log\dfrac{1}{\delta}\right\rceil \label{eqn:TheoreticalGuarantee.CMS.Row}
	\end{align}
	where $e$ is the base of the natural logarithm.

	CMS has a property that provides a theoretical bound of the estimated value.
	It satisfies

	\begin{equation}
		\label{eqn:TheoreticalGuarantee.ProbabilityEstimation}
		P(\hat c\le c+\varepsilon\|\hat c\|)\ge 1-\delta
	\end{equation}
	where $c$ is $c(e;\theta)$ or $c(g;\theta)$, $\hat c$ is the approximation of $c$, $\|\hat c\|$ is the sum of all elements in the CMS for $\hat c$, i.e.,
	\begin{equation}
		\|\hat c\|\coloneqq\sum_{i=1}^d\sum_{j=1}^w\hat c_{ij}
	\end{equation}

	At the same time, due to the overestimating nature of the CMS, it also satisfies
	\begin{equation}
		a\le\hat a
	\end{equation}
	where $a$ is $a(e;\theta)$ or $a(g;\theta)$, $\hat a$ is the approximation of $a$.

	Now, we define the adjusted approximation $\tilde c$.

	\begin{equation}
		\tilde c\coloneqq\hat c-\varepsilon\|\hat c\|
	\end{equation}
	and thus the adjusted G-test statistic.

	\begin{equation}
		\tilde G\coloneqq\left|2\tilde c\cdot\log\left[\dfrac{\tilde c\cdot(t-1)}{\hat a}\right]\right|
	\end{equation}

	With these, we can give the theoretical bound of the adjusted G-test statistic.

	\begin{theorem}
		\label{thm:TheoreticalGuarantee}
		Let $\chi^2_{1-\delta}(1)$ be the $1-\delta$ quantile of a $\chi^2$ random variable with 1 degree of freedom.
		Then
		\begin{equation}
			P(\tilde G>\chi^2_{1-\delta}(1))<\delta
		\end{equation}
		where $\delta$ is a parameter of the CMS;
		$\tilde G$ is the adjusted G-test statistic.
	\end{theorem}

	In other words, if $\tilde G$ is the test statistic and $\chi^2_{1-\delta}(1)$ is the threshold, the probability of producing a false positive, i.e., $\tilde G$ is incorrectly higher than $\chi^2_{1-\delta}(1)$, is at most $\delta$.

	\begin{proof}
		The distribution of $G$ is approximately a $\chi^2$ distribution with $k-1$ degrees of freedom, where $k$ is the number of classes.
		This also holds for our modified G-test statistic.
		In Isconna, there are two classes, i.e., accumulated and current, so the degree of freedom is 1.

		For the real G-test statistic, it satisfies

		\begin{equation}
			\label{eqn:TheoreticalGuarantee.ProbabilityChiSquared}
			P(G\le\chi^2_{1-\delta}(1))=1-\delta
		\end{equation}

		If we take the union bound of Eq.~\ref{eqn:TheoreticalGuarantee.ProbabilityEstimation} and Eq.~\ref{eqn:TheoreticalGuarantee.ProbabilityChiSquared}, with a probability of at least $1-\delta$, both events can occur.
		Then,

		\begin{align}
			\tilde G & =\left|2\tilde c\cdot\log\left[\dfrac{\tilde c\cdot(t-1)}{\hat a}\right]\right|                         \\
			         & =\left|2(\hat c-\varepsilon\|\hat c\|)\log\left[\dfrac{(\hat c-\varepsilon\|\hat c\|)(t-1)}{\hat a}\right]\right| \\
			         & \le\left|2c\cdot\log\left[\dfrac{c\cdot(t-1)}{a}\right]\right|                                          \\
			         & =G\le\chi^2_{1-\delta}(1)
		\end{align}

		Eventually, we can conclude that

		\begin{equation}
			P(\tilde G>\chi^2_{1-\delta}(1))<\delta
		\end{equation}
	\end{proof}

	\section{Dataset Details}

	The statistical summary of the dataset used in the experiments is shown in Table~\ref{tab:Dataset}.
	The time resolution of all datasets is one second.
	For simplicity, timestamps are shifted to start from 1.

	\begin{table*}[!htb]
		\centering
		\caption{Summary of datasets}\label{tab:Dataset}
		\begin{tabular}{llrrr}
			\toprule
			Dataset   & Source                           & \# nodes & \# edge records & \# timestamps \\
			\midrule
			CIC-DDoS2019 & \cite{Sharafaldin2019Developing} & 1,290    & 20,364,525      & 12,224        \\
			CIC-IDS2018  & \cite{Sharafaldin2018Generating}     & 33,176   & 7,948,748       & 38,478        \\
			CTU-13       & \cite{Garcia2014empirical}       & 371,100  & 2,521,286       & 33,090        \\
			DARPA        & \cite{Lippmann1999Results}       & 25,525   & 4,554,344       & 46,567        \\
			ISCX-IDS2012 & \cite{Shiravi2012developing}         & 30,917   & 1,097,070       & 165,043       \\
			UNSW-NB15    & \cite{Moustafa2015UNSW}          & 50       & 2,540,047       & 85,348        \\
			\bottomrule
		\end{tabular}
	\end{table*}

	Links and instructions below help readers to access the datasets used in experiments.

	\begin{itemize}
		\item CTU-13: https://mcfp.weebly.com/the-ctu-13-dataset-a-labeled-dataset-with-botnet-normal-and-background-traffic.html
		\item ISCX-IDS2012: https://www.unb.ca/cic/datasets/ids.html
		\item UNSW-NB15: https://www.unsw.adfa.edu.au/unsw-canberra-cyber/cybersecurity/ADFA-NB15-Datasets/
		\item CIC-DDoS2019: https://www.unb.ca/cic/datasets/ddos-2019.html
		\item DARPA: https://www.ll.mit.edu/r-d/datasets/1998-darpa-intrusion-detection-evaluation-dataset
		\item CIC-IDS2018: Run aws s3 sync --no-sign-request "s3://cse-cic-ids2018/Processed Traffic Data for ML Algorithms/" . in AWS CLI. Note the dot at the end represents the current directory.
	\end{itemize}

	\section{Parameter Search}\label{sec:Appendix.ParameterSearch}

	We performed parameter searches to obtain the best performance for baselines and our method.
	The tested parameter values are based on the default values provided by the original authors.
	We try to make sure our search range effectively covers all the typical values.
	However, due to the slow running speed of the python implementation, for F-Fade and PENminer, the number of tested combinations is limited.

	\paragraph{SedanSpot}

	\begin{itemize}
		\item \texttt{sample\_size} $\in\{500,2000,10000\}$
		\item \texttt{num\_walk} $\in\{10,50,200\}$
		\item \texttt{restart\_prob} $\in\{0.15,0.5\}$
	\end{itemize}

	\paragraph{PENminer}

	\begin{itemize}
		\item \texttt{ws} $=1$
		\item \texttt{ms} $=1$
		\item \texttt{view} = \texttt{id}
		\item \texttt{alpha} $=1$
		\item \texttt{beta} $\in\{0.2,1\}$
		\item \texttt{gamma} $\in\{1,5\}$
	\end{itemize}

	\paragraph{F-Fade}

	For \texttt{t\_setup}, we always use the timestamp value at the 10th percentile of the dataset.
	Other tweaked parameters are \texttt{W\_upd} $\in\{120,360,720\}$ and \texttt{T\_th} $\in\{60,120\}$.

	\paragraph{MIDAS}

	The size of CMSs is 2 rows by 3000 columns for all the tests.
	For MIDAS-R, the decay factor $\alpha\in\{0.3,0.5,0.7\}$.

	\paragraph{Isconna}

	The size of CMSs is 2 rows by 3000 columns for all the tests.
	The decay factor $\zeta\in\{0,0.3,0.5,0.7\}$, the frequency weight $\alpha=1$.
	For the width weight $\beta=0$, only the gap weight $\gamma=0$ is tested;
	for $\beta=1$, $\gamma\in\{0.5,1\}$ are tested.

	\section{Best Parameters of Isconna}\label{sec:Appendix.BestParameter}

	Table~\ref{tab:Appendix.BestParameter} gives the parameters used to produce the highest AUROC on each dataset.
	The corresponding AUROC are reported in the main article.

	\begin{table}[!htb]
		\centering
		\caption{Best parameters of Isconna}
		\label{tab:Appendix.BestParameter}
		\begin{tabular}{lcllllll}
			\toprule
			Dataset      & Variant & $\alpha$ & $\beta$ & $\gamma$ & $\zeta$ & $r$ & $c$  \\
			\midrule
			CIC-DDoS2019 & EO      & 1        & 1       & 0.5      & 0.7     & 2   & 3000 \\
			CIC-DDoS2019 & EN      & 1        & 1       & 0.5      & 0.7     & 2   & 3000 \\
			CIC-IDS2018  & EO      & 1        & 1       & 0.5      & 0.7     & 2   & 3000 \\
			CIC-IDS2018  & EN      & 1        & 0       & 0        & 0.7     & 2   & 3000 \\
			CTU-13       & EO      & 1        & 1       & 1        & 0.7     & 2   & 3000 \\
			CTU-13       & EN      & 1        & 0       & 0        & 0.5     & 2   & 3000 \\
			DARPA        & EO      & 1        & 0       & 0        & 0.7     & 2   & 3000 \\
			DARPA        & EN      & 1        & 0       & 0        & 0.7     & 2   & 3000 \\
			ISCX-IDS2012 & EO      & 1        & 1       & 0.5      & 0.7     & 2   & 3000 \\
			ISCX-IDS2012 & EN      & 1        & 1       & 0.5      & 0.7     & 2   & 3000 \\
			UNSW-NB15    & EO      & 1        & 1       & 1        & 0.7     & 2   & 3000 \\
			UNSW-NB15    & EN      & 1        & 1       & 1        & 0.7     & 2   & 3000 \\
			\bottomrule
		\end{tabular}
	\end{table}

	\bibliography{main}